\documentclass[runningheads]{llncs}
\usepackage{eccv}

\usepackage{eccvabbrv}

\usepackage{graphicx}
\usepackage{booktabs}

\usepackage[accsupp]{axessibility} 

\usepackage{multirow}
\usepackage[table]{xcolor}
\usepackage{array}
\usepackage{colortbl}
\usepackage{xcolor} 

\usepackage[ruled,vlined,linesnumbered]{algorithm2e}
\SetKwInput{KwInput}{Input}
\SetKwInput{KwOutput}{Output}
\SetKwInput{KwRequire}{Require}
\SetKw{KwRet}{return}
\DontPrintSemicolon
\SetAlFnt{\small}

\usepackage{hyperref}

\usepackage{orcidlink}

\begin{document}

\title{AutoSpeed: Annotation-Free Stage-Adaptive Motion Speed Learning for Robot Manipulation} 

\titlerunning{AutoSpeed}

\author{Qingda Hu\inst{1}\textsuperscript{*} \and
Ziheng Qiu\inst{1}\textsuperscript{*} \and
Jieru Zhao\inst{2}\and
Zhongxue Gan\inst{1}\and
Wenchao Ding\inst{1}\textsuperscript{\dag}
}

\authorrunning{Q.~Hu et al.}

\institute{College of Intelligent Robotics and Advanced Manufacturing, Fudan University, Shanghai, China \and
School of Computing, Shanghai Jiao Tong University, Shanghai, China
\\[0.5em] \textsuperscript{*}Equal contribution. \quad
\textsuperscript{\dag}Corresponding author: \email{dingwenchao@fudan.edu.cn}
}

\maketitle

\begin{abstract}
Different stages of manipulation tasks exhibit varying levels of difficulty, suggesting stage-dependent motion speeds and temporal prediction horizons.
However, existing IL-based visuomotor policies typically imitate the execution speed of expert demonstrations and operate with a fixed temporal prediction horizon, limiting flexibility and overall task throughput.
In this paper, we introduce \textbf{AutoSpeed}, a model-agnostic learning framework that enables existing visuomotor policies to predict trajectories with stage-adaptive motion speeds, without requiring speed or stage annotations.
We treat future trajectories at different speeds as candidate optimization targets, evaluate each candidate using a composite cost that trades off prediction error against prediction horizon, and optimize the policy toward the minimum-cost candidate.
With a fixed-length action sequence, speed modulation adjusts the effective temporal prediction horizon: simple stages are executed faster with a longer prediction horizon, whereas complex stages are executed more slowly with a shorter prediction horizon.
Specifically, we implement speed modulation in the frequency domain via the discrete cosine transform (DCT), which enables smooth, non-integer speed scaling and thus preserves motion continuity.
Extensive evaluations show that AutoSpeed substantially reduces task execution time while also improving success rates. Under the AutoSpeed framework, the inferred motion speeds exhibit a strong correspondence with task stages.

\end{abstract}

\begin{figure}[!t]
    \centering
    \includegraphics[width=\textwidth]{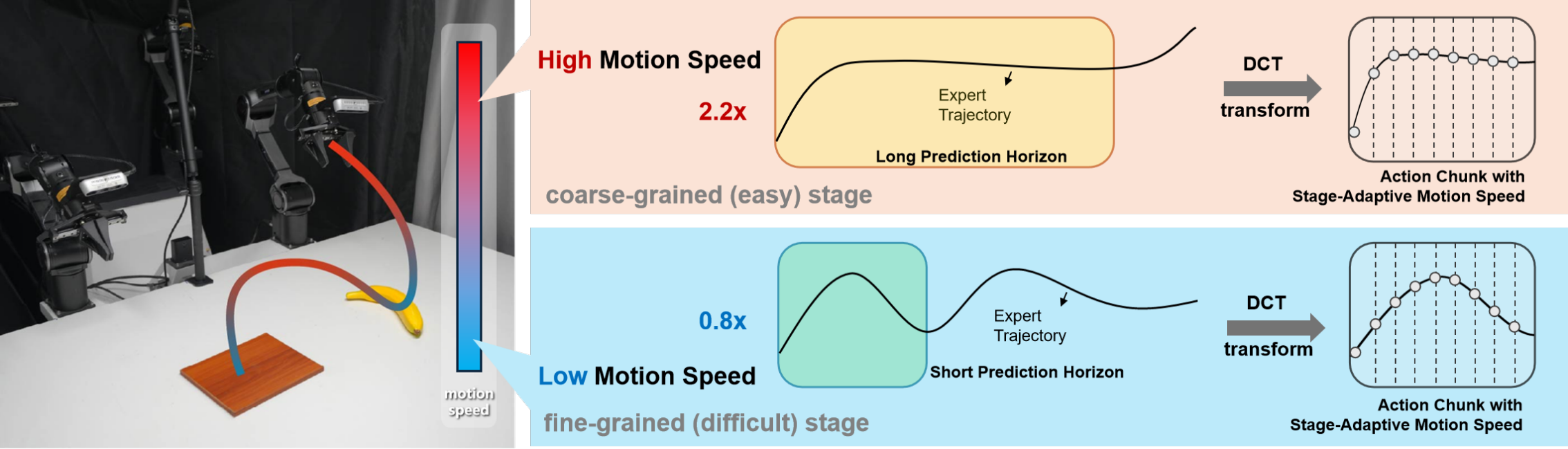}
    \caption{Stage-aware motion speed adaptation. Motion speeds in expert demonstrations are often suboptimal. AutoSpeed aims to train policies to predict future trajectories with stage-aware motion speed without requiring speed or stage annotations. 
}
    \label{fig:coarse}
\end{figure}

\section{Introduction}
\label{sec:intro}

Imitation learning (IL) is widely adopted for visuomotor policy learning. In recent years, it has catalyzed a diverse range of visuomotor policies~\cite{scaledp,dp,baku,uva} for robotic manipulation, including prominent Vision-Language-Action (VLA) models~\cite{octo,pi0,rdt,dreamvla}.
Existing IL-based policies typically mimic the motion speed exhibited in expert demonstrations. \cite{better-than-il-2019}
Regardless of the stage of the task, the policy predicts actions over a fixed future horizon, i.e., it uses fixed-length action chunking~\cite{act,dp}. 
However, the motion speeds in expert demonstrations are often suboptimal, thereby limiting the efficiency and performance of the learned policy.
In practical deployment scenarios, task completion efficiency is often critical, especially in industrial settings.

Different stages of manipulation tasks exhibit varying levels of difficulty~\cite{stage_aware1,stage_aware_reward}, suggesting that both the motion speed and the temporal prediction horizon should be stage-dependent.
Empirically, we find a consistent positive relationship between motion speed and effective prediction horizon, aligning with prior findings in human motor and cognitive control~\cite{human1,human2}.
As shown in Fig.~\ref{fig:coarse}, in easy stages (e.g., free-space reaching, or gross repositioning), long-horizon action chunks can be predicted reliably from the current observation~\cite{bid,chunkselect}, allowing faster execution. In contrast, fine-grained stages (e.g., insertion, or sustained interaction) demand higher precision and tighter perception--action coupling, and are therefore better served by shorter effective prediction horizons and slower execution.
This is consistent with findings in human motor and cognitive control: when task demands are low, humans can act faster with relatively little monitoring, whereas precision-critical behaviors require closer feedback monitoring and often slower movements to preserve accuracy ~\cite{heuer2023speedaccuracy,badre2025cognitivecontrol,pierrieau2025flexibility}.

Building on these motivations, we introduce AutoSpeed, a model-agnostic  learning framework that enables existing visuomotor policies to predict trajectories at stage-adaptive motion speeds.
In this work, we hypothesize and validate that the stage-dependent speed ratio can be implicitly inferred during end-to-end policy training. This obviates the need to train an additional proxy policy~\cite{demospeedup} or to rely on VLM-based or manual annotations~\cite{espada_2512_entropy,BFA,longvla}. Moreover, AutoSpeed jointly optimizes the quality of implicit speed ratio inference and trajectory prediction in an end-to-end training pipeline.

Concretely, as shown in Fig.~\ref{fig:coarse}, given a set of speed ratios (e.g., sampled from 0.8 to 2.2 in increments of 0.2), we obtain action chunks at different speeds by applying discrete cosine transform (DCT)\cite{dct-theory} to the original trajectories in the frequency domain. This technique enables non-integer speed modulation while preserving more high-frequency action details for fine-grained manipulation\cite{FreqPolicy}. 
We use fixed-length action chunks to keep the model’s output dimensionality consistent; acceleration compresses trajectories in time and increases the effective horizon, while deceleration expands them and reduces it.
We then treat future trajectories at different speeds as a set of candidate supervision targets. Each candidate is evaluated with a composite cost that trades off prediction error against prediction horizon, and we optimize the policy toward the minimum-cost candidate (section~\ref{sec:3.2optimize}). 
A lightweight Ratio Head is trained jointly to predict the speed ratio from latent observation features. Additionally, similar to the temporal ensemble~\cite{act}, we introduce a nonlinear temporal ensemble (section~\ref{sec:nta}) that outputs actions by combining the speed prediction of the ratio head with multiple overlapping fragment predictions, allowing for smooth deployment at inference time.

Through extensive experiments across diverse simulation benchmarks and real-world tasks, we show that visuomotor policies trained with AutoSpeed consistently reduce task execution time while improving success rates. AutoSpeed supports both single-task and multi-task learning and works across policy classes, spanning non-generative (e.g., MLP-based) and generative (e.g., diffusion- and flow-based) action prediction models. We further demonstrate additional capabilities of AutoSpeed, including controllable motion style and improved policy performance on datasets with substantial demonstration-speed variability.
We summarize the contributions of this paper as follows:
\begin{itemize}
    \item[1.] We introduce AutoSpeed, which to our knowledge is the first annotation-free, stage-aware framework to implicitly infer motion speed ratios within end-to-end policy learning.
    \item[2.] With DCT-based frequency-domain scaling, AutoSpeed enables non-integer acceleration and deceleration while preserving high-frequency action details.
    \item[3.] Policies trained with AutoSpeed substantially shorten task completion time while improving success rates. Moreover, the inferred speed ratios are closely aligned with task stages.
\end{itemize}

\section{Related Work}
\subsection{Motion Speed Modulation in Robot Manipulation}
We categorize the literature on motion speed modulation in robot manipulation by when the motion speed ratio is inferred: (1) pre-training inference, where it is inferred before training and used to annotate the dataset; (2) test-time inference, where it is inferred during model deployment; and (3) training-time emergence from unlabeled data, where it is implicitly inferred during end-to-end training.

The following methods fall into the first category:
DemoSpeedUp~\cite{demospeedup} derives safe-to-accelerate regions by estimating action entropy with a proxy policy, and retrains a faster policy on the annotated dataset. Similarly, ESPADA~\cite{espada_2512_entropy} uses VLM/LLM-based stage annotations to identify the safe-to-accelerate regions. Moreover,
SpeedAug~\cite{speedaug} constructs tempo-augmented trajectories to learn a tempo-enriched imitation prior, then applies RL-based fine-tuning to push toward faster policies. 
And the following methods fall into the second category:
SAIL~\cite{sail} integrates complexity-aware speed modulation with high-gain tracking and scheduling to safely increase execution speed, while SRIL~\cite{SRIL_2410_skipaction} reduces computing latency by learning when to skip inference. 
VFIL~\cite{VFIL_2410} treats speed as an explicit command and conditions the policy on sampling frequency, enabling variable-speed execution by changing the control frequency at deployment without retraining. 

However, the third paradigm remains underexplored despite its practical importance. Annotation-based approaches face two key limitations. First, introducing external tools or training additional models to produce annotations is costly and often requires task-specific heuristics or rules. Second, annotation quality and cross-trajectory consistency directly affect the final policy performance, yet these annotations are not optimized end-to-end with the policy. 
Moreover, prior work largely focuses on when to accelerate; in AutoSpeed, we unify acceleration and deceleration within a single speed-modulation perspective.

\subsection{Flexible Reconfiguration of Action Chunking}

Action chunking, which models the joint distribution of future actions conditioned on past states, is a standard practice in robotic manipulation~\cite{act,dp}. While action chunking strengthens temporal consistency and mitigates error accumulation, long chunks reduce access to the most recent observations during execution and thus limit reactivity~\cite{bid}.
A growing body of work shows that the prediction horizon in action chunking can substantially impact policy performance, suggesting the existence of an optimal horizon~\cite{autohorizon,bid,rtc}.
Embodied manipulation is inherently stage-structured: predictability and feedback requirements vary substantially across stages, so a single global optimal horizon is at best an average-case compromise. A more appropriate view is local optimality, where the effective horizon should be adapted to the current context during policy prediction or execution.

From the training-time perspective, 
MoH~\cite{mixhorizon} formulates action chunking as a mixture over multiple horizons, processes them in parallel with a shared action transformer, and fuses the outputs via a gating module.
From the inference-time perspective, 
AutoHorizon~\cite{autohorizon} adjusts the execution horizon online based on the model’s attention weights.
GBC~\cite{selfguidechunkadaptive} improves diffusion-based behavior cloning via self-guidance adaptive chunking mechanism that selectively refreshes action chunks when higher reactivity is needed.

In contrast to these horizon-adaption designs, AutoSpeed does not explicitly enumerate or select action chunks with different horizons.
Instead, the policy is directly optimized toward the optimal action target, avoiding the need to model suboptimal horizon modes or incur the inference-time cost of grouped prediction.
Moreover, the speed ratio emerges implicitly during policy learning, making the mechanism model-agnostic and easy to apply across policy architectures.

The shared motivation between motion speed modulation and action chunking reconfiguration is to move beyond a single globally tuned trade-off toward stage-aware, locally optimal adaptation across task stages.
Dynamic adjustment of the prediction horizon covered by action chunks is also one of our motivations, and we find it is strongly positively correlated with motion-speed modulation across task stages.
Accordingly, we keep the output chunk size fixed to preserve architectural compatibility, and use motion-speed modulation to induce a coupled change in the effective inference horizon, enabling stage-adaptive temporal reasoning without altering the model output dimensionality.

\begin{figure}[!t]
    \centering
    \includegraphics[width=\textwidth]{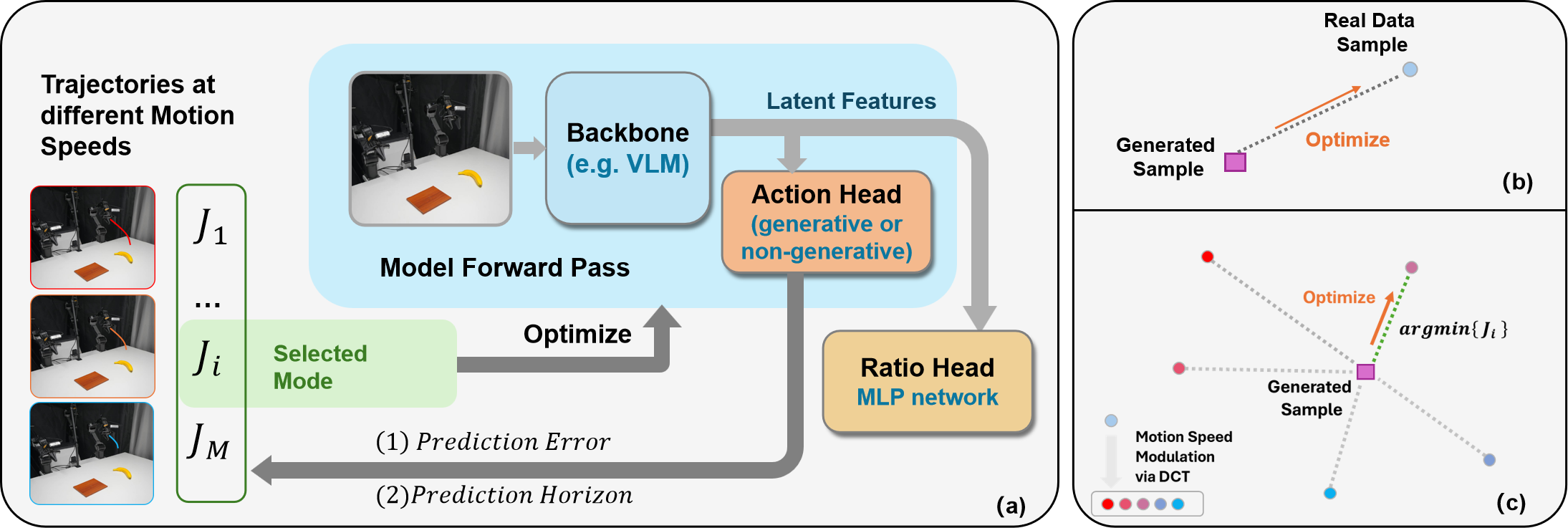}
    \caption{Overview of AutoSpeed. Training with AutoSpeed is formulated as a cost-aware multi-target selective optimization problem, as illustrated in (a), where trajectories at different motion speeds form a set of candidate supervision targets. AutoSpeed performs mode selection by minimizing a composite cost $J$ that trades off prediction error against prediction horizon, and optimizes the policy toward the minimum-cost target. (b) and (c) contrast generative-model training with and without AutoSpeed. With AutoSpeed (c), each sample is guided toward the target that attains the lowest cost. }
    \label{fig:pipi}
\end{figure}

\section{Method}
\label{method}

\subsection{Problem Setup}
\label{Sec:3.1}
Imitation learning for robotic manipulation aims to train a policy by minimizing the prediction error of demonstrated trajectories conditioned on observations.
An expert dataset is given as $\mathcal{D}=\{(l^{(i)}, \tau^{(i)})\}_{i=1}^{N},$, where each trajectory $\tau$ is denoted by
\(
\tau=\{(o_t, a_t, s_t)\}_{t=1}^{T}
\) and the task instruction $ l $ specifies the task.
At each time step \(t\), \(o_t\) denotes the image observation, \(a_t\) denotes the motion control signal, and \(s_t\) denotes the robot state. We denote the chunk size by $H$, and the observation horizon by $K$. For brevity, we omit conditioning inputs other than $\mathbf{o}_{t-K+1:t}$. A standard visuomotor policy \(\pi_\theta\) predicts a length-\(H\) action chunk conditioned on the recent observation context:
\begin{equation}
\pi_\theta\!\left(\mathbf{a}_{t:t+H-1}\mid \mathbf{o}_{t-K+1:t}\right).
\end{equation}

To enable stage-adaptive motion speed learning, the key challenge is to identify which stages in demonstration trajectories can be safely accelerated.
We define the motion speed ratio as $r_t$ at time step $t$.
Larger $r_t$ corresponds to easier stages, allowing faster execution and a longer temporal prediction horizon; smaller $r_t$ corresponds to precision-critical stages, demanding slower execution, a shorter horizon and higher-frequency feedback.

In our framework we use a fixed number of chunks to keep the model’s output tensor shape consistent. Therefore, the temporal prediction horizon $h_t$ is positively correlated with the motion speed ratio $r_t$. then:
\begin{equation}
\mathrm h_t \ =\;{H} \cdot r_t   \
\end{equation}

Accordingly, a visuomotor policy trained with AutoSpeed is formulated as:
\begin{equation}
\pi_\theta\!\left(\mathbf{a'}_{t:t+{h_t}-1},\mid \mathbf{o}_{t-K+1:t}\right).
\end{equation}

where $\mathbf{a'}_{t:t+h_t-1}$ has a motion speed $r_t$ matched to the current task stage.
The transformation from $\mathbf{a}_{t:t+H-1}$ to $\mathbf{a'}_{t:t+{h_t}-1}$ is described in Sec.~\ref{DCT}. Besides,
Sec.~\ref{sec:3.2optimize} details the AutoSpeed optimization procedure. Sec.~\ref{Sec3.4other} presents training recipes for generative policy classes and introduces a compatible inference-time strategy that aggregates overlapping predictions to improve smoothness. In the following sections, we denote an action chunk by $A$.

\subsection{Multi-Target Selective Optimization}
\label{sec:3.2optimize}

Recent studies have analyzed and empirically validated quantifiable stage signatures in robotic manipulation: fine-grained behaviors are often reflected in the high-frequency components of action trajectories\cite{FreqPolicy}, while the policy’s predicted action entropy can serve as a proxy for identifying difficult, failure-prone stages\cite{25nips_entropy_FIPER,demospeedup,rss25MIentropy}.
We hypothesize that the stage-dependent signal $r_t$ can be implicitly inferred during end-to-end training, and in this work we validate that it is feasible and model-agnostic. We formulate training as cost-aware multi-target optimization.

\subsubsection{Optimization Target Set}
At each time step $t$, we construct a set of candidate future trajectories at different speeds, denoted by
\(
\{A^{(m)}\}_{m=1}^M
\). 
AutoSpeed supports a user-defined discrete set of motion-speed ratios (e.g., sampled from 0.8 to 2.2 in increments of 0.2), and $M$ denotes the number of speed candidates. 
Compared to manually downsampling demonstrations at only integer-rate factors\cite{espada_2512_entropy,demospeedup}, this design provides greater flexibility and enables smoother speed transitions.

\subsubsection{Cost-based Selective Optimization}
As shown in Fig.~\ref{fig:pipi}, the multi-target setting yields a set of losses, each corresponding to an optimization direction. We evaluate these candidates using a composite cost $J$ that trades off prediction error against temporal prediction horizon:
\begin{equation}
J\left({A^{(m)}}\right)
=
\frac{
\mathcal{E}^{(m)}_t
}{
w + \log\!\left(h_t\right)
},
\label{eq:composite-cost}
\end{equation}

\(\mathcal{E}^{(m)}_t\) denotes the model's mean-squared prediction error (MSE) for candidate \(m\) at time \(t\).
For non-generative models, \(\mathcal{E}^{(m)}_t\) reduces to the MSE between the predicted action chunk and \(A^{(m)}\).
For generative models, \(\mathcal{E}^{(m)}_t\) can be instantiated as an MSE either on the denoising target (i.e., the injected noise) or directly on the ground-truth actions\cite{JiT}, with the former being more commonly used\cite{dp,rdt,flowpolicy}.
The denominator serves as an adaptive normalization term.
Candidates with shorter prediction horizons receive a larger penalty, encouraging the model to favor well-predicted futures while selecting faster speeds whenever possible. Intuitively, in easier stages, future actions remain predictable over longer horizons, permitting higher speed ratios; in precision-critical stages, long-horizon prediction becomes less predictable, yielding lower speed ratios.
Additionally, \(w\) is a constant term that can be tuned to control motion style. A smaller \(w\) biases the policy toward a more aggressive motion style.

We then select the minimum-cost candidate:

\begin{equation}
m^\star
=
\arg\min_{m\in\{1,\dots,M\}}
J\!\left({\mathbf{A}}^{(m)}\right)
\label{eq:target-selection}
\end{equation}

AutoSpeed trains the policy by minimizing the loss on the selected target,
\begin{equation}
\min_{\theta}\;
\mathbb{E}_{\tau\sim\mathcal{D}}
\left[
\sum_{t}
\mathcal{L}_\theta\!\left(\mathbf{o}_{t-K+1:t},\, \mathbf{A}^{m^{\star}}_{t}\right)
\right].
\label{eq:elasticmotion-objective}
\end{equation}

\subsubsection{Criterion for different models.}
AutoSpeed is model-agnostic. For policies with non-generative action heads,
the model output is the action prediction; therefore, we set $\mathcal{E}^{(m)}_t$ to the mean-squared error (MSE) of the predicted actions.

For diffusion-based action heads, the action chunk \(A^{(m)}\) serves as the generation target and is progressively corrupted by Gaussian noise \(\boldsymbol{\epsilon}\sim\mathcal{N}(\mathbf{0},\mathbf{I})\) during training to obtain a noised sample \(\mathbf{x}^{(m)}_\kappa\) at diffusion step \(\kappa\); the denoiser \(\epsilon_\theta(\cdot)\), conditioned on the observation context \(\mathbf{o}_{t-K+1:t}\), learns to predict the injected noise. Therefore, we set \(\mathcal{E}^{(m)}_t\) to the mean-squared error (MSE) of the predicted noise:
\begin{equation}
\mathcal{E}^{(m)}_t
=
\mathbb{E}_{\boldsymbol{\epsilon}\sim\mathcal{N}(\mathbf{0},\mathbf{I}),\,\kappa}
\left[
\left\|
\boldsymbol{\epsilon}-
\epsilon_\theta\!\left(\mathbf{o}_{t-K+1:t},\,\mathbf{x}^{(m)}_\kappa,\,\kappa\right)
\right\|_2^2
\right],
\label{eq:Em-diff}
\end{equation}

For flow-matching-based action heads, the model learns a conditional velocity field \(v_\theta(\cdot)\) that transports a noise sample \(\boldsymbol{\epsilon}\) to the target action chunk \(A^{(m)}\) along an interpolation \(\mathbf{x}^{(m)}_\tau=(1-\tau)\boldsymbol{\epsilon}+\tau A^{(m)}\), where \(\tau\in[0,1]\) is the interpolation time.
Accordingly, we set
\begin{equation}
\mathcal{E}^{(m)}_t
=
\mathbb{E}_{\tau\sim \mathcal{U}(0,1),\,\boldsymbol{\epsilon}\sim\mathcal{N}(\mathbf{0},\mathbf{I})}
\left[
\left\|
v_\theta\!\left(\mathbf{o}_{t-K+1:t},\,\mathbf{x}^{(m)}_\tau,\,\tau\right)
-
\left(A^{(m)}-\boldsymbol{\epsilon}\right)
\right\|_2^2
\right].
\label{eq:Em-flow}
\end{equation}

\subsubsection{Ratio Head.}
We introduce a lightweight Ratio Head trained to predict the speed ratio $r_t$ from latent observation features. 
It provides stage-adaptive motion speeds in the third stage of generative-model training~\ref{sec-gen-recipi}, and its predictions are used by the Nonlinear Temporal Aggregation (NTA) module~\ref{sec:nta} during inference.

\subsection{Motion Speed Transform}
\label{DCT}
We apply the discrete cosine transform (DCT)~\cite{dct-theory} to map time-domain action signals into the frequency domain, perform temporal scaling there to realize stretching/compression along the time axis, and then transform the actions back to the time domain. 

In details, we first compute DCT-II coefficients of the original action chunk $A\in\mathbb{R}^{H_0\times D}$ along the temporal axis, where $H_0$ denotes the maximum number of action samples required:
\begin{equation}
\mathbf{C}=\mathrm{DCT}\!\left(A\right)\in\mathbb{R}^{H_0\times D}
\end{equation}

We then retime by evaluating the inverse-DCT basis at the speed ratio $r$ and reconstruct the retimed chunk via a basis--coefficient product:
\begin{equation}
\tilde{A}=\mathbf{B}(r)\mathbf{C},\qquad
\mathbf{B}(r)_{i,k}=\sqrt{\frac{2}{H_0}}\,\alpha_k\cos\!\Big(\frac{\pi k (t_i+0.5)}{H_0}\Big),
\end{equation}
where $t_i=i\cdot r$ for $i=0,\dots,H-1$, $k=0,\dots,H_0-1$, $\alpha_0=1/\sqrt{2}$, and $\alpha_k=1$ for $k\ge 1$.
During training, the candidates $\{\tilde{A}^{(m)}\}_{m=1}^{M}$ are generated by applying the operator with the speed ratio set $\{r_m\}_{m=1}^{M}$.

This enables both acceleration and deceleration with non-integer scaling factors, while the frequency-domain processing effectively preserves high-frequency details that are crucial for fine-grained manipulation\cite{FreqPolicy}.

\subsection{Other Techniques}
\label{Sec3.4other}

\subsubsection{Training Recipe for Generative Models}
\label{sec-gen-recipi}

For non-generative models, applying AutoSpeed introduces no additional model computation overhead. For generative models, AutoSpeed incurs a small additional computational cost: the model’s action head needs to denoise each noise-perturbed trajectory candidate. However, this overhead is limited for two reasons. First, the parameters of the action head typically constitute only a small fraction of the overall model\cite{pi0,baku,dreamvla}. Second, the denoising of different trajectory candidates can be performed in parallel\cite{li2018imle,GCP}, which prevents the training process from being significantly slowed down.

To further improve training stability for generative models and reduce the additional computational cost, we propose a three-stage training strategy:
Inspired by~\cite{25nips_whyDPnotMemo}, which shows that diffusion models tend to preferentially learn smooth, low-complexity, and generalizable structure in the early stages of training, we optimize the model in the early training stage by minimizing the average loss over all candidates.
The second stage corresponds to the multi-objective optimization process (Sec.~\ref{sec:3.2optimize}). In the third stage, the Ratio Head is frozen, and the model is optimized directly using the motion speed predicted by the Ratio Head, allowing the remaining steps in the generation process to rapidly converge to the selected mode.

\subsubsection{Nonlinear Temporal Aggregation}
\label{sec:nta}
Conventional temporal aggregation assumes uniform time intervals consecutive actions across all chunks, rendering it incompatible with the variable-speed predictions generated by AutoSpeed. To address this limitation, we introduce Nonlinear Temporal Aggregation (NTA), a novel mechanism that enables the ensembling of overlapping prediction across action chunks operating at different temporal scales.

NTA computes a weighted average of actions from multiple chunks within a predefined time window. Specifically, at a given control step $s$ (denoted temporally as $t=0$), all predicted actions from previously generated chunks that fall within the time window $[-i,+i]$ are collected. 
Then, we compute a weighted average of the candidate actions using an exponential-decay weighting scheme and slight temporal misalignment is also corrected via the same frequency-domain transformation.

\section{Evaluation}
\label{sec:exp}
Our evaluations aim to investigate and answer the following questions:
\begin{itemize}
    \item[1.] Can AutoSpeed effectively reduce the time required to complete tasks while maintaining task success rates?
    \item[2.] Can the inferred speed ratios produced by AutoSpeed effectively indicate different task stages?
    \item[3.] Is AutoSpeed effective in both single-task and multi-task learning settings, and does it work for both non-generative and generative models?
\end{itemize}

\begin{figure}[!t]
    \centering
    \includegraphics[width=\textwidth]{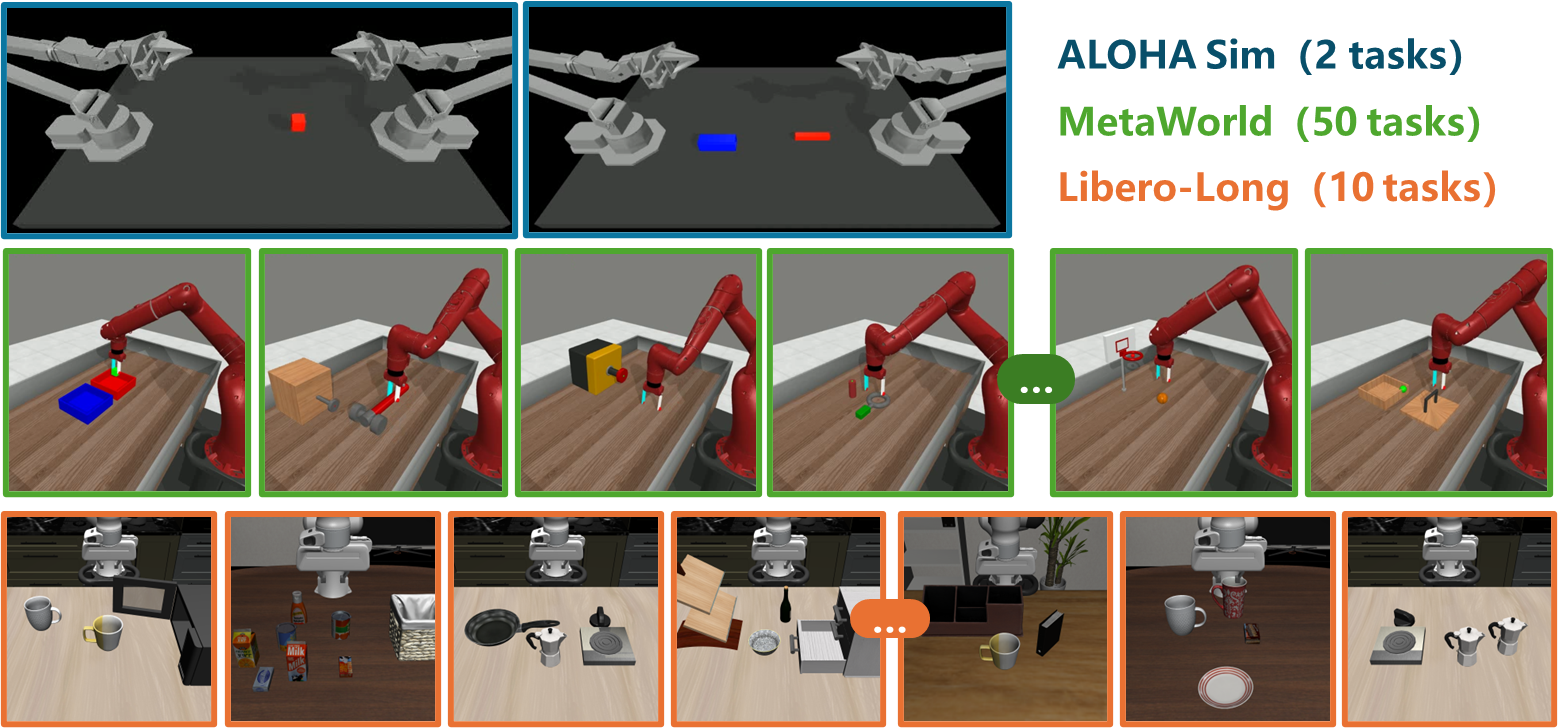}
    \caption{\textbf{Simulation Tasks.} We select a total of 62 tasks from ALOHA Simulation\cite{act}, MetaWorld\cite{yu2020metaworld}, and LIBERO-Long\cite{liu2023libero}, and use 50 demonstration trajectories for each task.}
    \label{fig:simtasks}
\end{figure}

\subsection{Experimental Setup}

We evaluate AutoSpeed across a range of standard manipulation benchmarks and real-world tasks. 

\subsubsection{Simulation Tasks:} We experiment with two bimanual tasks from the ALOHA benchmark~\cite{act}, 10 tasks from the LIBERO-10 suite~\cite{liu2023libero}, 50 tasks from the Meta-World benchmark~\cite{yu2020metaworld}, shown in Fig.~\ref{fig:simtasks}. In ALOHA simulation, we evaluate the Transfer Cube and Insertion tasks, utilizing 50 expert demonstrations per task. For both the LIBERO-10 suite and the Meta-World benchmark, the policies are trained using 50 expert demonstrations per task under a multi-task learning paradigm.

\subsubsection{Real-world Tasks:} We conduct real robot experiments on an Agilex Piper bimanual robot platform in a tabletop manipulation environment. The policies are trained on RGB images and robot proprioceptive state. The images are captured via two wrist cameras mounted on the top of the grippers and one fixed camera installed on top of the table. The action space comprises the joint states and the gripper state. Using a teleoperation system, we collect a dataset for four distinct tabletop tasks, with each task comprising approximately 50 expert demonstrations. Notably, the expert action trajectories are temporally oversampled by a factor of 2, enabling fine-grained actions when deploying deceleration.

\subsubsection{Models}
We compare policies optimized with our AutoSpeed framework against their vanilla counterparts trained with original objectives. 

\paragraph{ACT}: Action Chunking Transformer(ACT)~\cite{act} is a visuomotor policy composed of a transformer encoder and decoder. We use it as a representative non-generative baseline model. 

\paragraph{BAKU}: BAKU\cite{baku} is a transformer-based architecture designed for multi-task learning. This architecture is highly representative: most existing visuomotor policies, including VLA models, are largely built upon an Observation Encoder--Backbone--Action Head design.
It employs a transformer encoder to fuse information from different modalities and trains a decoupled action head to generate action chunks. This modular design enables the action head to be readily substituted with several generative variants.

\subsubsection{Metrics}
In addition to the standard metric of task success rate, we report the average episode length of successful rollouts to assess overall execution efficiency. For the Aloha benchmark, we perform 50 evaluation rollouts per task. For both the LIBERO-10 suite and the Meta-World benchmark, we adopt a consistent evaluation protocol, executing 10 trials per task and reporting the aggregated success rates and execution length. For the real-world evaluation, we report the task success rate and average completion time (in seconds) across 20 trials per task.

\begin{figure}[!t]
    \centering
    \includegraphics[width=\textwidth]{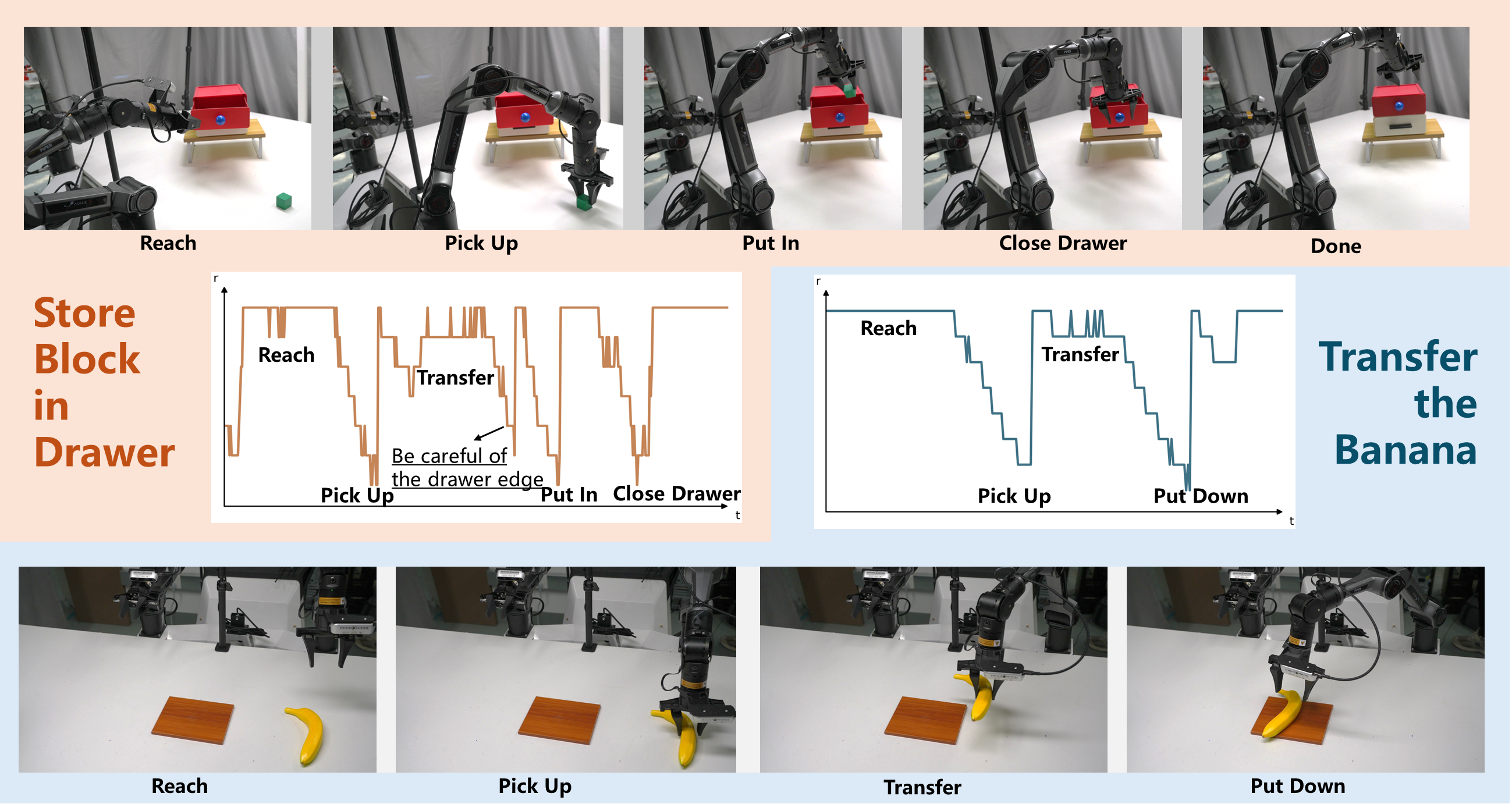}
    \caption{The speed ratio curves of two real-world tasks correspond to the task stages. Under the AutoSpeed framework, the inferred speed ratios closely align with task stages. }
    \label{fig:tasks}
\end{figure}

\subsection{Main Results on Single-Task Learning Simulation}

Table~\ref{tab:sim_results_single_task} summarizes the quantitative results evaluated on the ALOHA benchmark.

Compared to the vanilla ACT baseline, the policy trained with AutoSpeed (denoted as \textit{AutoSpeed}) achieves a substantial reduction in task execution time without compromising overall performance. 
Specifically, on the Transfer Cube task, AutoSpeed reduces the average episode length from 272 to 160 steps, yielding a maximum speedup of 1.7x. While this accelerated execution slightly lowers the success rate (from 72\% to 64\%) under standard control due to the heightened sensitivity of high-speed interactions, coupling AutoSpeed with a high-gain controller effectively mitigates this issue. This combination not only preserves the execution efficiency but also surpasses the original baseline with a 6\% improvement. Furthermore, on the contact-rich Insertion task, AutoSpeed simultaneously improves the success rate (from 22\% to 24\%) and reduces execution length (from 353 to 296 steps with high-gain), demonstrating its robust execution efficiency in single-task learning scenarios.

\begin{table*}[htbp] 
    \centering
    \caption{Quantitative results across multiple tasks on the ALOHA benchmark. $\dagger$ denotes evaluated with a high-gain controller.}
    \label{tab:sim_results_single_task}
    \begin{tabular}{lcccc} 
        \toprule
        \multirow{2}{*}{\textbf{Method}} & \multicolumn{2}{c}{\textbf{Transfer Cube}} & \multicolumn{2}{c}{\textbf{Insertion}} \\
        \cmidrule(lr){2-3} \cmidrule(lr){4-5} 
        & \textbf{SR} ($\uparrow$) & \textbf{Len} ($\downarrow$) & \textbf{SR} ($\uparrow$) & \textbf{Len} ($\downarrow$) \\
        \midrule
        ACT                   & 72\% & 272 & 22\% & 353 \\
        ACT(\textit{AutoSpeed})             & 64\% & \textbf{160} & \textbf{24\%} & 300 \\
        ACT(\textit{AutoSpeed}$^{\dagger}$) & \textbf{78\%} & 165 & \textbf{24\%} & \textbf{296} \\
        \bottomrule
    \end{tabular}
\end{table*}

\subsection{Main Results on Multi-Task Learning Simulation}
The quantitative results on the LIBERO-10 and Meta-World benchmark are presented in Table~\ref{tab:sim_results_multi_task}.  AutoSpeed consistently outperforms the strongest baselines across both multi-task benchmarks.
On Meta-World, AutoSpeed yields substantial improvements in success rates regardless of the underlying action head: BAKU-DiT exhibits a 19.0\% increases, while BAKU-Flow improves by 7.8\% to achieve the highest overall success rate (65\%). Crucially, these gains are accompanied by marked reductions in average episode lengths.
On the LIBERO-10 suite, AutoSpeed-trained policy compresses the execution time with a maximum acceleration of approximately 40\% to its vanilla counterpart.  Compared to baselines that naively execute actions at a fixed, dataset-dictated speed, AutoSpeed’s stage-adaptive optimization empowers policies to complete tasks substantially faster, while simultaneously improving success rates by dynamically adjusting to the complexity of the current task phase.

Crucially, our experiments demonstrate that reducing task execution time and improving success rates are not mutually exclusive. By predicting trajectories at stage-adaptive motion speeds, AutoSpeed leverages long-horizon action chunks for faster execution during easy stages, where future actions can be predicted reliably. Conversely, during fine-grained stages that demand tighter perception-action coupling, the framework naturally adopts slower execution and shorter effective horizons to preserve control accuracy. Because this stage-aware modulation is implicitly inferred during end-to-end training via our composite cost, AutoSpeed generalizes effectively across multi-task scenarios, delivering superior execution efficiency while preserving task success compared to baselines.

Beyond qualitative observations, we further analyze the consistency between the predicted speed ratios and action entropy from DemoSpeedUp~\cite{demospeedup}, which reflects divergence across multiple sampled action predictions, and find that unlike action entropy, which relies on grouped predictions from a proxy policy and struggles to distinguish uncertainty from multi-modality, AutoSpeed more accurately identifies safely acceleratable stages, such as robot start-up and return-to-home fast motions.

\begin{table*}[t]
\centering
\small
\setlength{\tabcolsep}{4pt}
\renewcommand{\arraystretch}{1.15}

\caption{Quantitative results on multi-task learning benchmarks. Success rate (SR, $\uparrow$) and episode length (Len, $\downarrow$).}
\label{tab:sim_results_multi_task}

\begin{tabular}{lcccc}
\toprule

\multirow{2}{*}{\textbf{Method}} 
& \multicolumn{2}{c}{\textbf{LIBERO-10}} 
& \multicolumn{2}{c}{\textbf{Meta-World}} \\

\cmidrule(lr){2-3} \cmidrule(lr){4-5}

& \textbf{SR} ($\uparrow$) 
& \textbf{Len} ($\downarrow$) 
& \textbf{SR} ($\uparrow$) 
& \textbf{Len} ($\downarrow$) \\

\midrule

BAKU-DiT          
& 50\% & 261 
& 43\% & 87 \\

BAKU-Flow        
& 47\% & 287 
& 57\% & 82 \\

BAKU-DiT (\textit{AutoSpeed})    
& 49\% & 259 
& 62\% & \textbf{70} \\

\rowcolor{gray!12}
BAKU-Flow (\textit{AutoSpeed})  
& \textbf{52\%} 
& \textbf{171} 
& \textbf{65\%} 
& 71 \\

\bottomrule
\end{tabular}
\end{table*}

\subsection{Main Results on Real-World Tasks}
As shown in Table~\ref{tab:real_results}, AutoSpeed 
consistently outperforms the baselines across all evaluated real-world tasks with various length and difficulty levels. 
Most notably, when integrated with our proposed Nonlinear Temporal Aggregation (NTA), the AutoSpeed-trained policy significantly reduces execution time, delivering an average speedup of approximately 1.78$\times$ across the diverse task suite.

Crucially, this substantial acceleration does not compromise manipulation precision but yields consistent absolute improvements in task success rates. As illustrated in Fig.~\ref{fig:tasks}, in the highly challenging \textit{Place the Toy} task, AutoSpeed increases the success rate by a notable 10\%. These physical deployments further validate AutoSpeed's capacity to autonomously extract stage-dependent motion speed patterns from unannotated demonstrations. By leveraging the proposed multi-target optimization, the policy translates these learned patterns into superior task efficiency and robust real-world performance.

Notably, AutoSpeed trained on non-oversampled data still outperforms the baselines, experiencing only a marginal performance drop compared to the oversampled version. Since action oversampling is practically feasible, it simply serves as an optional enhancement to further improve fine-grained action fidelity enabled by DCT-based retiming.

\begin{table}[t]
\centering
\small
\setlength{\tabcolsep}{4pt}
\renewcommand{\arraystretch}{1.15}
\caption{Real-world evaluation. Success rate (SR, $\uparrow$) and execution time in seconds (Time, $\downarrow$).}
\label{tab:real_results}

\begin{tabular}{lcccc}
\toprule
\multirow{2}{*}{\textbf{Method}} 
& \multicolumn{2}{c}{\textbf{Transfer the Banana}} 
& \multicolumn{2}{c}{\textbf{Place the Toy}} \\
\cmidrule(lr){2-3} \cmidrule(lr){4-5}
& \textbf{SR} ($\uparrow$) & \textbf{Time} ($\downarrow$) 
& \textbf{SR} ($\uparrow$) & \textbf{Time} ($\downarrow$) \\
\midrule
BAKU-Flow+TA 
& 77\% & 20.80s 
& 33\% & 18.13s \\
\rowcolor{gray!12}
BAKU-Flow (\textit{AutoSpeed})+NTA
& \textbf{83\%} & \textbf{11.24s}
& \textbf{43\%} & \textbf{10.13s} \\
\midrule
\multirow{2}{*}{\textbf{Method}} 
& \multicolumn{2}{c}{\textbf{Stacking Two Cubes}} 
& \multicolumn{2}{c}{\textbf{Store Block in Drawer}} \\
\cmidrule(lr){2-3} \cmidrule(lr){4-5}
& \textbf{SR} ($\uparrow$) & \textbf{Time} ($\downarrow$) 
& \textbf{SR} ($\uparrow$) & \textbf{Time} ($\downarrow$) \\
\midrule
BAKU-Flow+TA 
& 60\% & 50.11s 
& 77\% & 38.67s \\
\rowcolor{gray!12}
BAKU-Flow (\textit{AutoSpeed})+NTA
& \textbf{63\%} & \textbf{28.52s}
& \textbf{80\%} & \textbf{23.01s} \\
\bottomrule
\end{tabular}
\end{table}

\begin{figure}[!t]
    \centering
    \includegraphics[width=\textwidth]{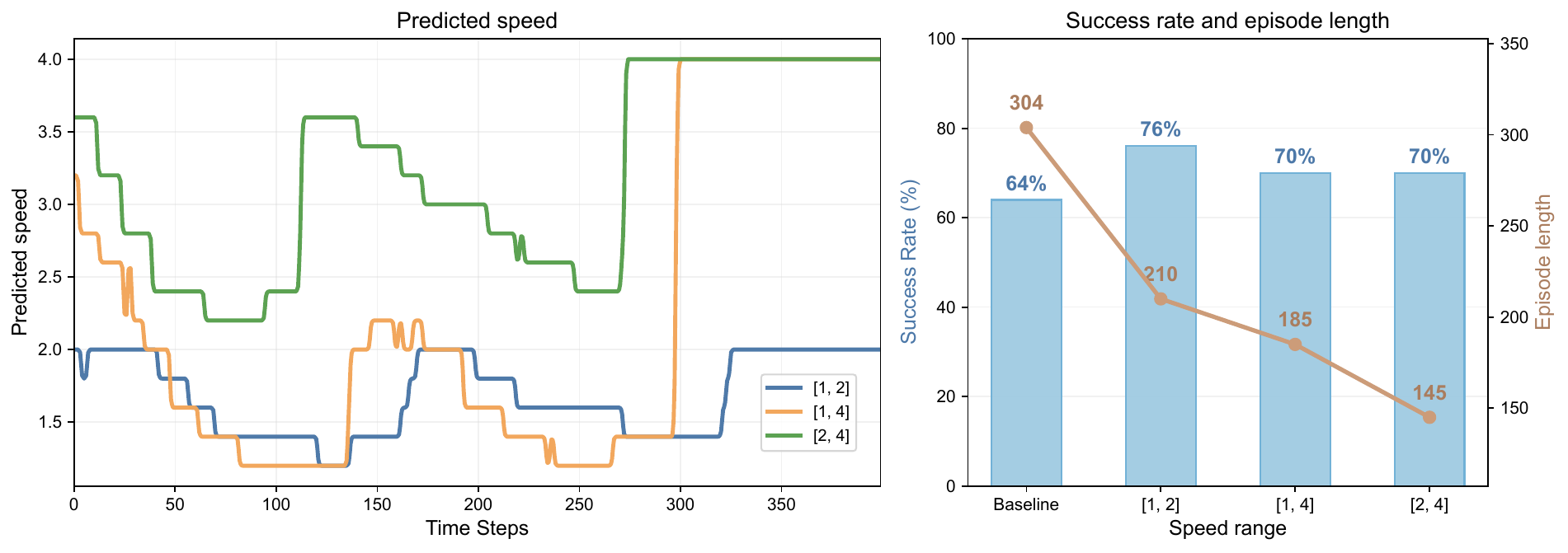}
    \caption{\textbf{Ablation on Speed Range Bounds.} \textbf{Left:} Predicted motion speed trajectories over time steps for AutoSpeed variants on the ALOHA Transfer Cube task. Despite different predefined bounds, all variants exhibit a consistent phase-aware pattern, autonomously decelerating during complex interaction stages. \textbf{Right:} Comparison of task success rates and average episode lengths with all variants outperforming the vanilla baseline.}
    \label{fig:ablation_speed}
\end{figure}

\subsection{Ablation Studies}

\subsubsection{Speed Range}
We evaluate how different predefined speed ranges impact the model's ability to learn phase-dependent motion patterns and its overall task performance. Specifically, we train the ACT policy equipped with AutoSpeed on the ALOHA Transfer Cube task under three distinct speed range settings: AutoSpeed [1x, 2x], AutoSpeed [1x, 4x], and AutoSpeed [2x, 4x], and a vanilla ACT model is included as the baseline. 
Note that while this ablation uses a reduced image resolution compared to the main experiments, the same resolution and all other hyperparameters are kept identical across all variants to ensure a fair comparison.

As illustrated in Fig.~\ref{fig:ablation_speed}(left), despite the varying upper and lower bounds, the learned speed trajectories across all three variants exhibit a remarkably consistent phase-aware pattern. The policies autonomously decelerate (forming a valley in the speed curve) during complex, interaction-critical task phases, and accelerate during simpler, unconstrained stages.

Furthermore, Fig.~\ref{fig:ablation_speed}(right) demonstrates that all three AutoSpeed variants consistently outperform the vanilla baseline in both success rate and execution efficiency. Notably, the results reveal a clear tradeoff dictated by the speed bounds: the more conservative [1x, 2x] setting achieves the highest success rate (76\% vs. the baseline's 64\%), while the aggressive [2x, 4x] setting maximally minimizes the episode length (dropping from 304 to 145 steps) while still maintaining a highly competitive success rate of 70\%.

\begin{figure}[!t]
    \centering
    \includegraphics[width=\textwidth]{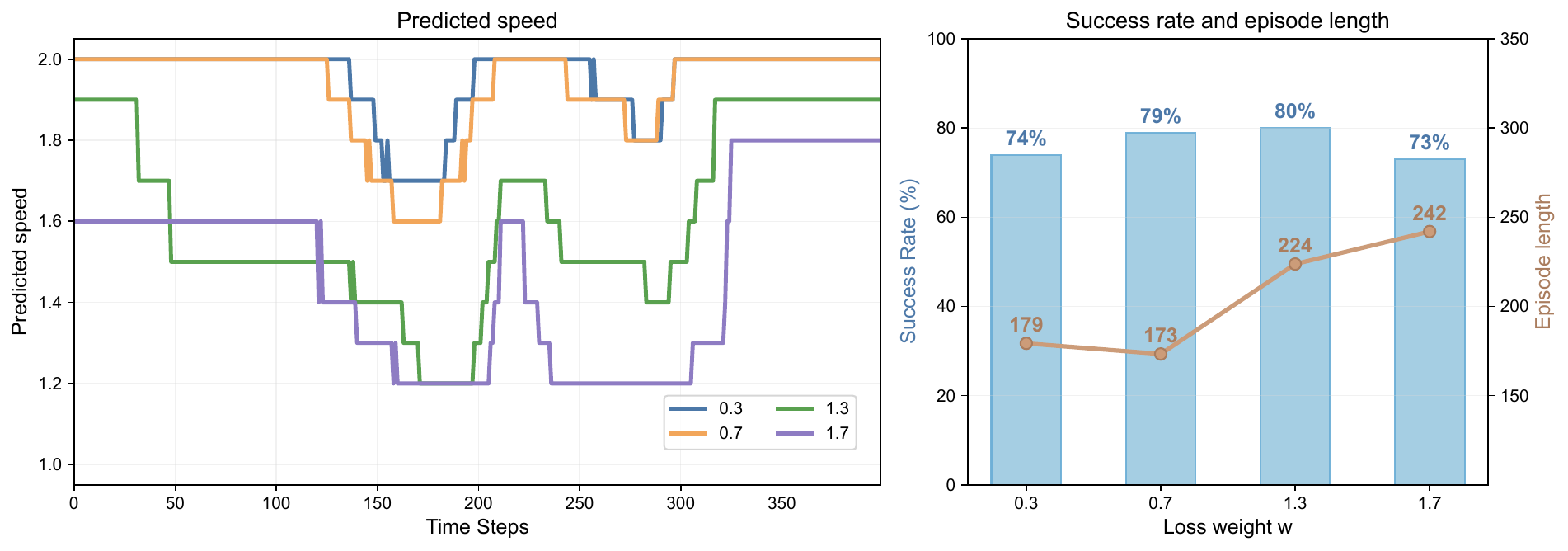}
    \caption{\textbf{Ablation on the length penalty coefficient $w$.} \textbf{Left:} Predicted motion speed trajectories over time steps for AutoSpeed variants on the ALOHA Transfer Cube task. \textbf{Right:} Comparison of task success rates and average episode lengths.}
    \label{fig:ablation_w}
\end{figure}

\subsubsection{Sensitivity to $w$}
\label{Sec:sensi_w}
The speed range and $w$ are both practical design choices for specific task requirements. 
Different length penalty coefficients inherently shift the model's preference for speed selection. As the penalty decreases, the relative cost of predicting shorter action chunk increases. Consequently, the model is penalized more heavily for selecting slower speeds and tends to predict action chunk with longer horizons. Fig.~\ref{fig:ablation_w} shows as policy trained with a smaller length penalty favors generating action chunk with higher speed. 
Overall, different choices share the phase-aware pattern: the policy accelerates in predictable stages and decelerates in critical interaction stages.

\section{Conclusion}
\label{sec:Conclusion}
AutoSpeed is a simple yet effective training paradigm for learning stage-adaptive motion speed and temporal prediction horizon directly from standard expert demonstrations. 
Rather than relying on external annotations or post-hoc trajectory retiming, AutoSpeed uncovers an endogenous signal \(r_t\) during end-to-end policy learning by selecting, among re-timed candidate futures, the target that best balances prediction error and temporal prediction horizon. 
This unifies speed control and temporal reasoning under a single objective and is compatible with both non-generative and generative visuomotor policies.
Across a broad suite of simulation and real-world manipulation tasks, AutoSpeed improves efficiency while maintaining success rates.
We hope AutoSpeed offers a practical and broadly applicable route toward faster, safer, and more reliable visuomotor policy deployment.

\bibliographystyle{splncs04}
\bibliography{main}

\end{document}